\documentclass{article}
\usepackage{amsmath,graphicx,mlspconf}
\usepackage{amssymb}
\usepackage{amsfonts}
\usepackage{algpseudocode,algorithm,algorithmicx}
\usepackage{subfigure}

\copyrightnotice{978-1-5090-6341-3/17/\$31.00 {\copyright}2017 IEEE}

\toappear{2017 IEEE International Workshop on Machine Learning for Signal Processing, Sept.\ 25--28, 2017, Tokyo, Japan}

\def\pa{{\text{\ttfamily pa}}}
\def\ch{{\text{\ttfamily ch}}}
\def\doi{{\text{\ttfamily do}}}
\def\T{{\text{T}}}
\def\E{{\mathbb{E}}}
\def\omegaB{{\boldsymbol \omega}}
\def\muB{{\boldsymbol \mu}}
\def\alphaB{{\boldsymbol \alpha}}
\def\xb{{\boldsymbol x}}

\def\B{{\boldsymbol B}}
\def\nb{{\boldsymbol n}}

\def\X{{\mathcal{X}}}
\def\V{{\mathcal{V}}}
\def\S{{\mathcal{S}}}

\title{Estimation of interventional effects of features on prediction}
\name{Patrick Bl{\"o}baum$^{1, 2}$, Shohei Shimizu$^{1,2,3}$\thanks{This work was supported by JSPS KAKENHI JP12345678 and ONRG NICOP N62909-17-1-2034.}}
\address{$^1$RIKEN AIP center, $^2$Osaka University, $^3$Shiga University, Japan}
\begin{document}
%

\maketitle
\begin{abstract}
The interpretability of prediction mechanisms with respect to the underlying prediction problem is often unclear. While several studies have focused on developing prediction models with meaningful parameters, the causal relationships between the predictors and the actual prediction have not been considered. Here, we connect the underlying causal structure of a data generation process and the causal structure of a prediction mechanism. To achieve this, we propose a framework that identifies the feature with the greatest causal influence on the prediction and estimates the necessary causal intervention of a feature such that a desired prediction is obtained. The general concept of the framework has no restrictions regarding data linearity; however, we focus on an implementation for linear data here. The framework applicability is evaluated using artificial data and demonstrated using real-world data.
\end{abstract}
\section{Introduction}
The predictive capabilities of machine learning methods have improved dramatically in recent years, primarily due to the availability of big data, increased computational speed, and recent breakthroughs in deep learning. However, although considerable focus is being placed on further improvement of prediction capabilities, deep understanding and interpretability of the prediction mechanisms are still lacking. For example, for the simplest setting of a linear prediction model, the meanings of the estimated coefficients remain mostly unclear with respect to the data generation process. A high coefficient of a predictor may indicate that this feature is particularly important for the prediction; however, the relationship between this coefficient and the actual underlying problem is not well understood. While studies on feature selection have investigated which features are important for the prediction, most have ignored the underlying causal relationship between the predictors and target \cite{Aliferis:2010:LCM:1756006.1756013}.

An interesting problem arises if we consider how to manipulate a data generation process such that a certain prediction can be expected. For example, if we have a classifier that classifies a patient as healthy or sick based on several observations such as a certain medicine's  dose, blood pressure, and heart rate, which feature must we change so that the patient is classified as healthy? If we have a classifier with a high coefficient for the feature representing a certain medicine's  dose, we could naively think that an increase/decrease of this dose may yield the classification of a healthy patient. Generally, this kind of interpretation of the prediction mechanism can have fatal consequences. An increase in the dose may cause an increase in the heart rate and blood pressure, which could further harm the patient. Instead, we must consider the causal relationship between the features. Therefore, a feature with a high coefficient in the prediction model may not necessarily be the best choice for manipulation in order to heal the patient. Further, if we apply feature selection, we may even remove important factors that have a significant causal influence on the chance of a cure.

In this paper, we draw a connection between a given prediction mechanism and the causal relationship of the predictors and target. We propose a framework consisting of three parts: 1) Integration of the causal structure of the prediction model into the causal structure of the underlying data generation process; 2) Identification of the feature with the greatest causal influence on the prediction; 3) Estimation of the necessary manipulation of a specific feature to achieve a certain prediction. To implement this framework, we consider linear causal relationships and linear predictions. For non-linear problems, the general concept of the framework remains unchanged, but modifications are required to account for the non-linear effects.

\vspace{-0.01cm}
\section{Background}
In the following, we introduce the notation and background theory. Generally, we denote a random variable $X$ by a capital letter, a specific value $x$ by a small letter, a set of random variables $\V = \{X_1, ..., X_i, ..., X_n\}$ by calligraphic letters, a vector $\xb = [x_1, ..., x_n]^\T$ by a small bold letter, and a matrix $\B$ by a capital bold letter.

\subsection{Causal graphical models}
\begin{figure}[t]\subfigure[]{
  \centering
  \includegraphics[width=0.31\columnwidth]{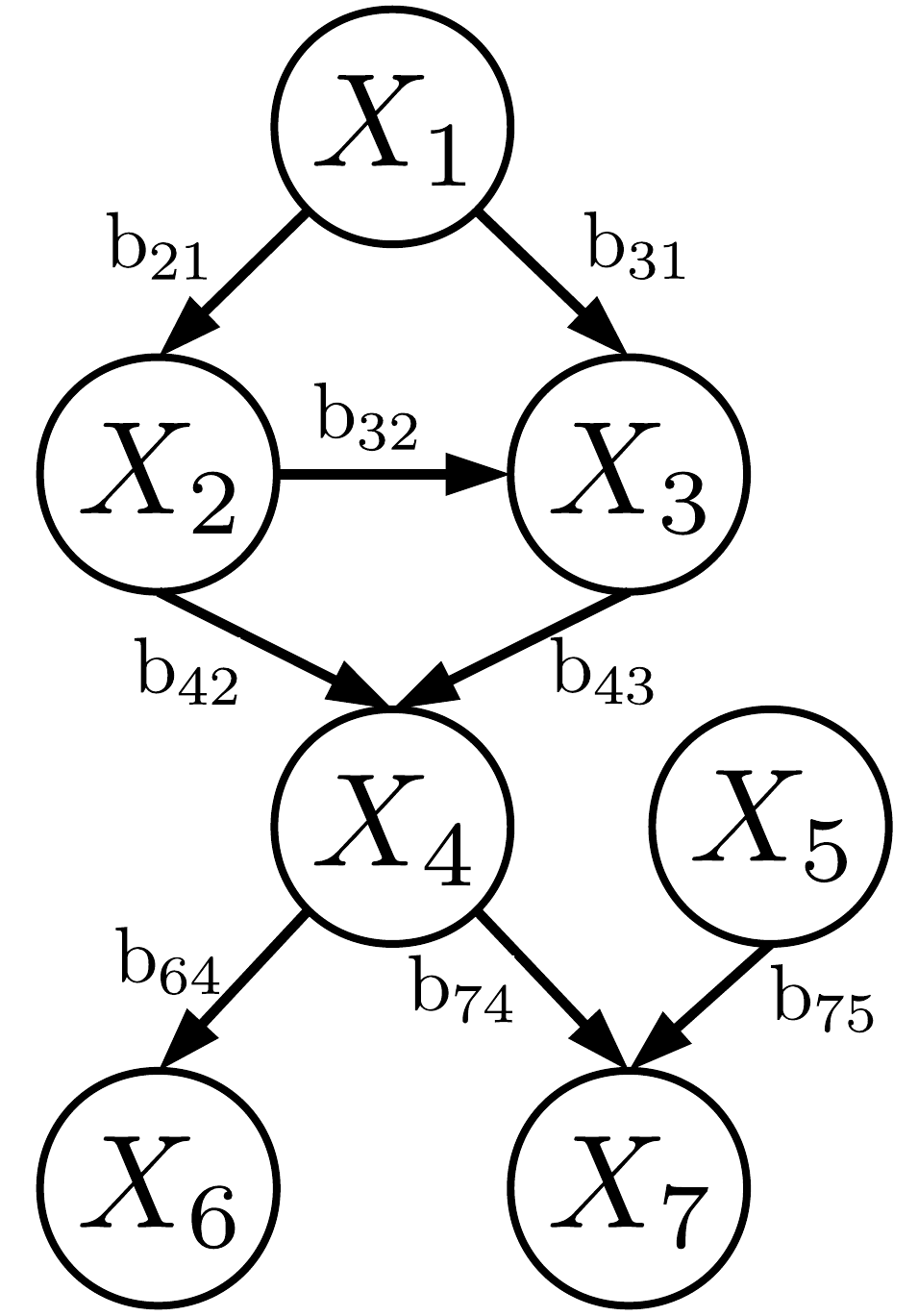}
  \label{fig:exampleDAG}
}
\subfigure[]{
  \centering
  \includegraphics[width=0.62\columnwidth]{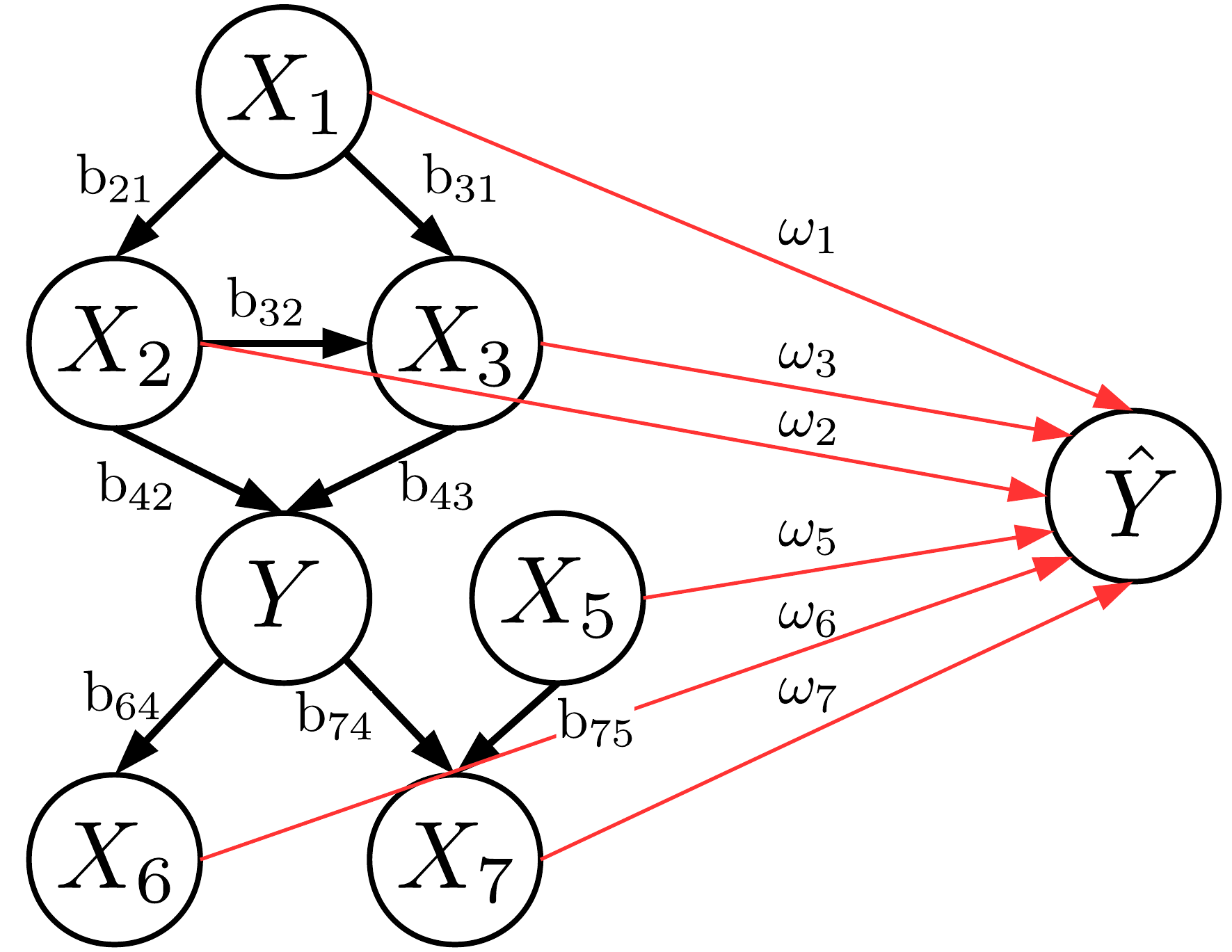}
  \label{fig:dagPred}
}
\caption{\textbf{(a)} Sample causally directed acyclic graph with seven variables. Each causal connection has a specific strength $b_{ij}$. \textbf{(b)} Combination of causal structure of prediction mechanism with the graph describing the data generation process. The target variable is $Y = X_4$. A new variable $\hat Y$ is introduced, where the red vertices indicate that all predictors are parents of $\hat Y$. The connection strengths are defined by the model coefficients $\omegaB$.}
\end{figure}

The causal structure of a set of variables $\V$ can be represented by a directed acyclic graph (DAG) \cite{Pearl:2009:CMR:1642718}, where each variable is a vertex. The direct causal influence of a variable $X_i$ on a variable $X_j$ is indicated by an arrow between these two vertices. Each connection has a certain strength that specifies the strength of the influence of $X_i$ on $X_j$. The entire causal structure can, therefore, be represented by a matrix $\B \in \mathbb{R}^{n \times n}$, where $b_{ij}$ represents the connection strength of $X_j$ to $X_i$. If $b_{ij} = 0$, no direct causal relationship exists. An example with seven variables is illustrated in Figure \ref{fig:exampleDAG}. The sets of parent and child variables of $X_i$ are denoted by $\pa(X_i)$ and $\ch(X_i)$, respectively, and include all directly connected ancestors and descendants of $X_i$, respectively. A variable with no ancestors is called a \textit{root} variable.

We assume that $X_i$ is affected by unobserved noise $N_i$. Therefore, in the linear case, the value of $X_4$ in Figure \ref{fig:exampleDAG}, for example, is defined by $X_4 = b_{42} \cdot X_2 + b_{43} \cdot X_3 + N_4$.

\subsection{Linear regression and classification}
As we restrict our implementation to linear predictions in this paper, we briefly discuss linear regression and linear classification. Generally, we assume that there exists one target variable $Y \in \V$ and we use all remaining variables as predictors $\X = \V \setminus \{Y\}$. However,  application of our framework to different arbitrary combinations of certain numbers of target or predictor variables is straightforward. For simplicity, we consider binary classification problems, which can also be easily extended to multiple categories.

For regression problems, we consider the prediction model
\begin{equation}
	\label{eq:linearReg}
	\hat Y = \phi(\X) = \omega_0 + \sum \limits_{X_i \in \X} \omega_i X_i,
\end{equation}
where $\omega_0$ and $\omega_i$ are the bias and the regression coefficient of predictor $X_i$, respectively. These coefficients can be represented by a vector $\omegaB$.

For classification problems, we utilize logistic regression, which can be formalized in a similar way as \eqref{eq:linearReg}, such that
\begin{equation}
	\label{eq:linearClass}
	\varphi(\X) = \ln \left(\frac{P(Y = 1 | \X)}{1 - P(Y = 0 | \X)} \right) = \omega_0 + \sum \limits_{X_i \in \X}  \omega_i X_i,
\end{equation}
where each $\omega_i$ represents the degree of influence of a predictor variable on the relative risk of being in class $1$ compared to the risk of being in class $0$. The classification rule is then defined as 
\begin{equation*}
	\hat y = \begin{cases}
0, \text{ if $\varphi(\xb) < 0$},\\
1, \text{ if $\varphi(\xb) > 0$},\\
\text{unknown otherwise}.
\end{cases}
\end{equation*}
If $\varphi(\xb) = 0$, we can force a decision by randomly choosing a label. A property of logistic regression is that the classification confidence of an observation $\xb$ increases with an increase in $|\varphi(\xb)|$. Note that, although we use the same notation $\omegaB$ for both the regression and classification model, it is clear from the context whether $\omegaB$ represents the linear or logistic regression coefficients.

Considering \eqref{eq:linearReg} and \eqref{eq:linearClass} and for a given observation $\xb$, we can easily estimate the appropriate value for $x_i$ such that we obtain the prediction $d$ by solving
\begin{equation}
	\label{eq:closedForm}
	x_i = \frac{d - \sum_{j \neq i}  \omega_j x_j - w_0}{w_i}.
\end{equation}

\subsection{Interventions and causal effect}
\label{sec:intervention}
Given a causal graph with variables $\V$, causal graphical models allow description of the manner in which the variables in $\V$ change if a variable $X_i \in \V$ is fixed to a certain value $c$. For example, in Figure \ref{fig:exampleDAG}, if we fix the value of $X_1$ to $c$, the changes in the remaining variables can be described by the causal connections defined by $\B$. In order to analyze this behavior, we first define the term \textit{intervention} as $\doi(X_i = c)$, which represents the fixing of $X_i$ to a constant value $c$. Note that, in the general context of causality, it is assumed that the causal structure and parameters remain unchanged under any kind of intervention.

The expected intervention effect of $X_i$ on another variable $X_j$ is called the \textit{total effect} \cite{Pearl:2009:CMR:1642718} and is expressed as
\begin{equation}
	\label{eq:interExp}
	\E[X_j | \doi(X_i = c)],
\end{equation}
which is the conditional expectation of $X_j$ given the intervention $\doi(X_i = c)$. For linear causal relationships, a further formalization and an algorithm for estimating \eqref{eq:interExp} for all variables are provided in Section \ref{sec:optimalInter}.

If we are only interested in the extent to which the intervention on $X_i$ affects $X_j$, we can estimate this influence via the derivative $\frac{\partial}{\partial X_i} \E[X_j | \doi(X_i = c)]$. In the linear case, this coincides with the regression coefficient $\omega^j_i$ of regressing $X_j$ on $X_i$ and $\pa(X_i)$ \cite{Pearl:2009:CMR:1642718}, where
\begin{align}
	\label{eq:causalEffect}
	X_j = \omega_0 + \omega^j_i X_i + \omegaB_{\pa(X_i)}^T \pa(X_i).
\end{align}
These regression coefficients can be obtained by minimizing the squared error. The coefficient $\omega^j_i$ is also called the \textit{causal effect} or the \textit{intervention effect} of $X_i$ on $X_j$. Therefore, the larger $|\omega^j_i|$, the smaller the changes in $X_i$ necessary to change $X_j$.

With respect to \eqref{eq:closedForm}, a naive intervention such as $\doi(X_i = (d - \sum_{j \neq i}  \omega_j x_j - w_0) / w_i)$ would also causally influence all descendants of $X_i$, yielding a different prediction of a post-intervention sample as the expected value $d$. Therefore, in our proposed framework, we take this causal influence into account in order to find the optimal $c$ for the intervention.

\section{Related work}
The causal structure of a given problem is not incorporated in most prediction applications. However, once we have a problem setting in which the population distribution changes, the causal structure can give valuable information regarding the implications of the prediction. Ref. \cite{peters2016causal} indicates that, because of the incorporation of causal information, the predictions of causally motivated prediction models are generally invariant under interventions and manipulations of variables. 
Related to this finding, Ref. \cite{tillman2011causality} reports the analysis of a causally motivated Markov blanket, where it was found that a prediction model is invariant under changes of variables that are not in the set of $\ch(Y)$. For improved causal interpretability and incorporation of causal information, Ref. \cite{2017arXiv170202604T} provides a causal regularization that can be applied to neural networks. The basic concept behind this regularization is the selection of variables that have a high likelihood of being causally related to the target and that are also significantly predictive. 

As regards the present work, we do not aim to develop a new prediction model or to achieve improved accuracy. Seeing that the outcome of the manipulations of certain processes or variables in a real-world problem often remain unclear with respect to the effect on the prediction, we rather want to provide new insights toward improved understanding of the relationship between a prediction problem and the corresponding prediction mechanism. Therefore, we focus on the actual prediction and analyze the causal influence of features on this prediction; this is achieved by combining the causal structure of the underlying data generation process with the causal structure of the prediction mechanism.

\section{Framework}
Our proposed framework consists of three steps:
\begin{enumerate}
	\item Training of a prediction model $f$ with predictors $\X$ and target $Y$ and integration of this model into the causal structure of the data generation process, which is defined by the set of variables $\V$ and causal connection matrix $\B$;
	\item Identification of the feature $X_i \in \X$ having the greatest intervention effect on the prediction $\hat Y = f(\X)$;
	\item Estimation of the intervention $\doi(X_i = c)$ such that the expectation of the prediction $\E[\hat Y | \doi(X_i = c)]$ of the post-intervention observations is equal or close to a specified value.
\end{enumerate}
In the third step, we essentially want to solve \eqref{eq:closedForm}, but considering the causal influence of a corresponding intervention. Note that we are not necessarily required to use all variables in $\X$ for the prediction. That is, this general procedure can be applied to any subset of $\X$, as all variables in $\V$ are considered in order to estimate the optimal intervention in the second and third steps.

\subsection{Causal structures of prediction mechanisms}
Let $Y \in \V$ denote the target variable we want to predict. If we consider the underlying causal structure of the discriminative prediction mechanisms defined by \eqref{eq:linearReg} and \eqref{eq:linearClass}, a clear causal relationship exists, where all variables in $\X$ are parents of the prediction $\hat Y$. However, if we examine Figure \ref{fig:exampleDAG}, for example, the causal structure of the prediction mechanism does not represent the real causal relationship between the predictors and target, except when we use $\pa(Y)$ as predictors only. In particular, as most prediction models are only designed to yield accurate predictions, $\hat Y$ may have a very different causal relationship with its predictors compared to the actual target $Y$ in the data generation process. For instance, the features in a Markov blanket of $Y$ are sufficient for an accurate prediction, but any descendant of $Y$ would be a parent of $\hat Y$ in the causal structure of the prediction. Further, in certain problems, we may want to make predictions according to a specific optimum criterion. Therefore, we can interpret the prediction models as a new data generation process that generates an output $\hat Y \neq Y$ based on $\X$. If we wish to combine the causal structure of $\V$ defined by $\B$ with the causal structure of the prediction mechanism, we must introduce a new variable $\hat Y$ with $\pa(\hat Y) = \X$ and connection strengths defined by the coefficients of the prediction model, as illustrated in Figure \ref{fig:dagPred}. The bias $\omega_0$ can be seen as as a constant noise affecting $\hat Y$.

\subsection{Estimation of optimal intervention}
\label{sec:optimalInter}
\begin{algorithm}[t]
\caption{Estimation of optimal linear intervention value.}
\begin{algorithmic}
\Function{GetInterventionValue}{$\muB, \B, \nb, \omegaB, y, i, d$}
\State $\S \gets \{1, ..., n\} \setminus \{i\}$
\State $\S \gets \text{RemoveIndicesOfRootVariables($\B$)}$
\State $\alphaB \gets \text{VectorOfZerosForEachVariable($\B$)}$
\State $\alpha_i \gets 1$
\State $\mu_i, n_i \gets 0$
\While{$\S$ is not empty}
	\State $k \gets \text{GetNextIndex($\S$)}$
	\If{$X_k$ has no parents in $S$}
		\State $\mu, \alpha \gets 0$
		\For{all parents of $X_k$}
			\State $q \gets \text{GetIndexOfNextParent($X_k$)}$
			\State $\alpha \gets \alpha + b_{kq} \cdot \alpha_q$
			\If{$q$ is not $i$}
				\State $\mu \gets \mu + b_{kq} \cdot \mu_q$
			\EndIf
		\EndFor		
		\State $\mu_k \gets \mu + n_k$
		\State $\alpha_k \gets \alpha$
		\State $\S \gets \text{RemoveIndex($k$)}$
	\EndIf
\EndWhile
\State $\omegaB \gets [\omega_1, ..., \omega_{y - 1}, 0, \omega_{y}, ..., \omega_{n - 1}]^\T$
\State $c \gets \frac{d - \omegaB^T \cdot \muB - \omega_0}{\omegaB^T \cdot \alphaB}$ \Comment Eq. \eqref{eq:closedForm}
\State \Return $c$
\EndFunction{}
\end{algorithmic}
\end{algorithm}

In Section \ref{sec:intervention}, we introduced the conditional expectation \eqref{eq:interExp} of a variable $X_j$ if an intervention $\doi(X_i = c)$ is performed. After integrating the structure of the prediction mechanism into the structure of the data generation process, we can now also investigate the manner in which the intervention of a variable affects the prediction and, further, determine what kind of intervention is necessary for a specific prediction.

In order to estimate $\E[\hat Y | \doi(X_i = c)]$, we must propagate through the causal graph and estimate the total effects of the intervention on all variables. For linear causal relationships, we can formalize this process as
\begin{align*}
	\E[X_j & | \doi(X_i = c)] = \\
	& \begin{cases}
	c, \text{ if $j = i$,}\\
\E[X_j], \text{ if $X_j$ is a root variable,}\\
\sum \limits_{X_k \in \pa(X_j)} \omega_k \E[X_k | \doi(X_i = c)] + \E[N_j], \text{ otherwise},
\end{cases}
\end{align*}
where $X_k$ denotes the $k$-th parent variable of $X_j$ and $\omega_k$ is the corresponding regression coefficient of regressing $X_j$ on $\pa(X_j)$. Note that $\omega_k$ is equal to the connection strength between $X_k$ and $X_j$ in $\B$. For non-linear relationships, we must integrate over all parents \cite{Pearl:2009:CMR:1642718}.

For our framework, we are interested in the intervention $\doi(X_i = c)$ such that, in the expectation, the prediction model predicts a value $d$. As this cannot be represented in a generic closed form solution such as that given in \eqref{eq:closedForm}, we formalize this problem as a minimization problem, with
\begin{equation}
	\label{eq:minProblem}
	c = \underset{c^*}{\operatorname{arg \ min}} \ (\E[\hat Y | \doi(X_i = c^*) ] - d)^2.
\end{equation}
Here, $c$ is the optimal intervention on $X_i$ that minimizes the squared difference between the actual post-intervention expectation of the prediction and the desired expectation. Note that we formalize this problem as a minimization problem in order to acquire a more general formulation of the problem that facilitates the easy addition of further constraints. Further, this allows utilization of arbitrary linear or non-linear models for $f$. Here, again, it is important to note that $\E[\hat Y | \doi(X_i = c) ] \neq \E[Y | \doi(X_i = c) ]$, if $\X \neq \pa(Y)$.  Therefore, the intervention is optimal with respect to $\hat Y$ in the original population and not with respect to $Y$. However, to find the optimal intervention with respect to $Y$, it is simply necessary to set $\X = \pa(Y)$ as predictors.

In our implementation, we restrict $f$ to represent either the linear regression $\phi$ or a logistic regression $\varphi$. In the latter case, we recall that $|d|$ implicitly represents the confidence of the classification in logistic regression. Therefore, by controlling the parameter $|d|$, it is possible to find an intervention such that we can expect the post-intervention observations to be assigned to a certain class with an arbitrarily high likelihood. 

We provide a pseudo code in Algorithm 1 that finds an exact solution of \eqref{eq:minProblem} in linear data without further constraints. The general concept behind the algorithm is to propagate through the causal network while monitoring the extent by which each variable would be affected by an intervention on variable $X_i$; this is represented by the values in vector $\alphaB$. Thus, the expectation of $X_j$ depends on the expectation of the parents, which again depends on the extent by which they are affected by a possible intervention. The parameters for this algorithm are as follows:

$\muB = [\E[X_1], ..., \E[X_n]]^T:$ A vector with the expectations of each variable before the intervention;

$\B$: The matrix with all causal connection strengths; this describes the causal structure;

$\nb = [\E[N_1], ..., \E[N_n]]^T$: A vector with the expectations of each noise variable;

$\omegaB$: The coefficients of the prediction model. For the algorithm, it does not matter whether this represents the coefficients of linear or logistic regression;

$y$: The index of the target variable;

$i$: The index of the variable to which we want to apply the intervention;

$d$: The desired expected post-intervention output of the prediction.

The noise expectations can be estimated by
\begin{equation}
	\label{eq:estNoise}
	\E[N_i] = \E[X_i] - \sum \limits_{X_k \in \pa(X_i)} \omega_k \E[X_k],
\end{equation}
where, again, $w_k$ represents the connection strength between $X_k$ and $X_i$ in $\B$. If $X_i$ is a root variable, the noise can generally be interpreted as either $N_i = X_i$ or $N_i \equiv 0$. However, our algorithm does not consider the noise values of the root variables. Note that we do not introduce any constraints on the values of each variable. Therefore, the optimal intervention may yield values outside the domains of some variables. We leave this point open for future work.

For some problems, we may rather be interested in obtaining the optimal intervention with respect to a specific observation. For example, if we know certain details concerning a patient, such as their age, if they smoke, or if a given disease is common within their family, we can simply replace the values in $\muB$ with these specific details, instead of the expectation over many observations. The same applies to the noise vector $\nb$, where we must only replace the expectations of the parents in \eqref{eq:estNoise} with the specific values.

\subsection{Entire linear framework}
To combine all steps of the framework, it is necessary to first learn $\omegaB$, identify $X_i$ with the greatest causal effect on $\hat Y$ according to \eqref{eq:causalEffect}, estimate $\muB$, estimate $\nb$ according to \eqref{eq:estNoise}, and apply Algorithm 1 to estimate $c$. Modifications for different problem settings, such as a certain subset of predictors or specific observations, are straightforward. Although we assume that the causal structure $\B$ is already known, if $\B$ is unknown, causal inference methods can be used to infer causal relationships in the observational data. For linear causal relations, the Linear Non-Gaussian Acyclic Model (LiNGAM) \cite{shimizu2006linear}, for example, could be utilized to estimate $\B$ under the assumption of non-Gaussian independent noise. In our framework, we justify the decision to intervene on the predictor with the greatest causal effect on $\hat Y$ by noting that this predictor would require the smallest change in order to change the prediction. However, any other variable could also be chosen for the intervention. Note that an intervention may affect $\hat Y$, but may not necessarily make physical sense in terms of manipulating $Y$. For instance, if $Y$ is a root variable, any intervention on predictors can affect $\hat Y$, but not $Y$.

For non-linear data, the algorithm of the framework must be modified in order to incorporate the non-linear prediction model and non-linear causal effects. However, seeing that, for example, a linear combination of basis functions could be used, these modifications are conceptually the same as in the linear case. 

\section{Experiments}
We evaluated our framework using artificial linear data and demonstrated its applicability using real-world data.

\subsection{Artificial data}
We generated random DAGs according to the generation process described in \cite{Janzing:2012:IAI:2169485.2170008}, considering linear relationships only. Generally, each DAG consisted of 20 root and 50 descendant variables. Because of space constraints, we omit further details on the data generation process here, but more details can be found in \cite{Janzing:2012:IAI:2169485.2170008}.

The problem setting of classification in our framework is equivalent to the problem setting of regression as, in logistic regression, we aim to find an optimal intervention such that the post-intervention expectation of the logistic regression in \eqref{eq:linearClass} is equal to a certain value $d$. Therefore, we focused our evaluations using artificial data on classification problems only. 

The classes in the training data were generated by choosing one of the 70 variables at random, which then served as target variable $Y$. Then, class $0$ or $1$ was assigned if $y_i \leq \operatorname{median}[Y]$ or $y_i > \operatorname{median}[Y]$, respectively. In our experiments, the goal was to estimate the necessary intervention on the variable with the greatest causal effect on the prediction, such that that post-intervention observations generated from the same DAG were likely to belong to class $1$ according to the classification model. Note that, as these interventions depend on the logistic regression model, where we obtain $\E[\varphi(\X) | \doi(X_i = c)] = d$, we can control the likelihood of a post-intervention sample belonging to class $1$ by changing $d$. In this manner, we also have a direct quality measurement; that is, we can count the number of post-intervention samples classified as class $1$. Note that, because of the median, both classes are balanced in the expectation before the intervention.

As the data generation process of each dataset was known, it was possible to generate further samples that followed the same causal structure and noise distributions. In this manner, we could evaluate the estimated intervention by keeping the according intervention variable constant and generate new post-intervention samples for all other variables.

In order to train the parameters for logistic regression, we used the Matlab implementation of logistic regression. We compared the optimal intervention value estimated by our framework according to \eqref{eq:minProblem} with the naive estimated intervention value according to \eqref{eq:closedForm}, which does not consider the causal effects. In both the optimal case and the naive case, we applied an intervention to the variable with the greatest causal effect on the prediction. For each evaluation, we generated $1000$ DAGs with each $1000$ training and post-intervention samples. 

\begin{figure}[t]
	\centering
 \includegraphics[width=1\columnwidth]{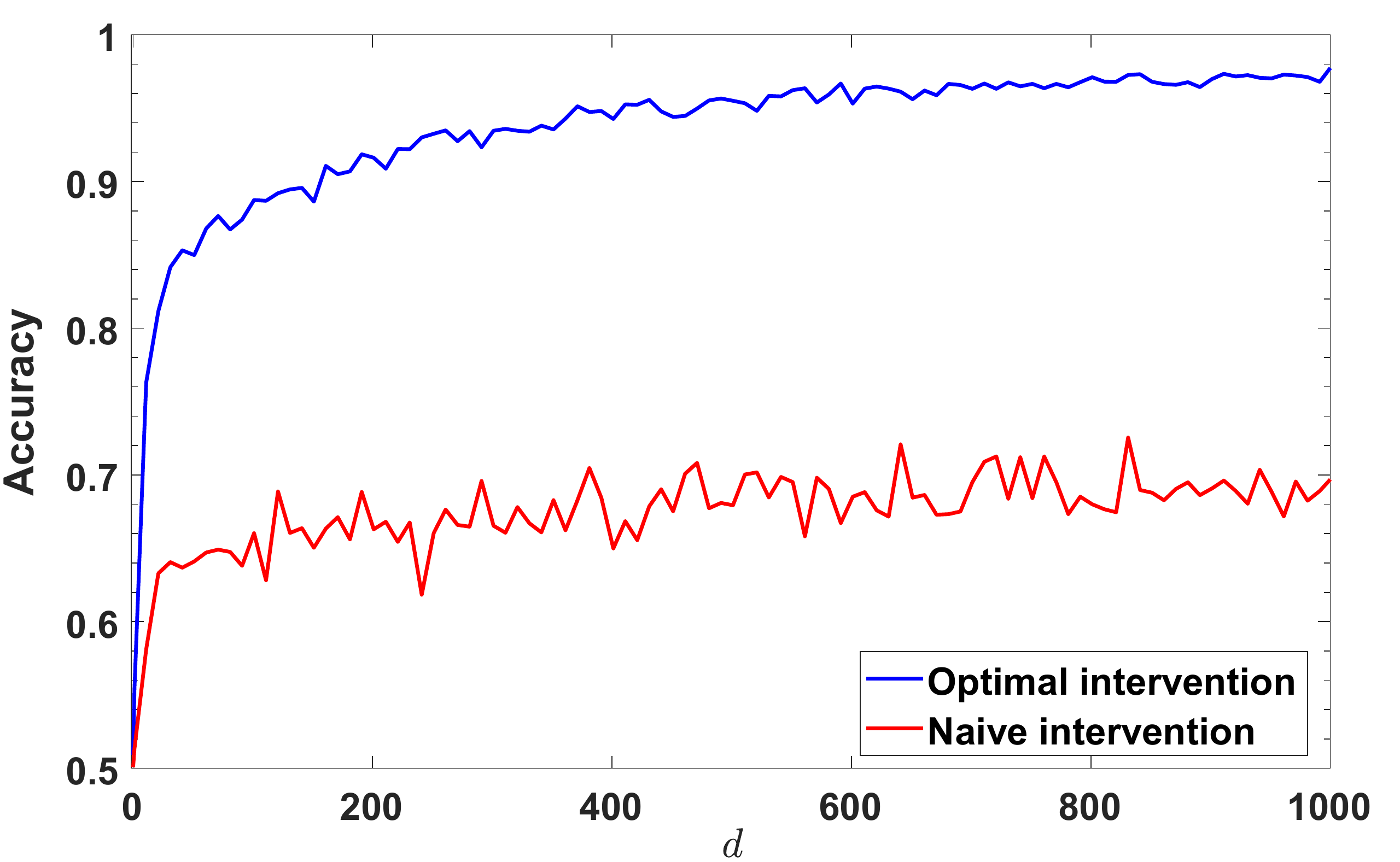}
\vspace*{-0.1cm}
\caption{Results of evaluations for classification problems with artificially generated DAGs. The accuracy corresponds to the ratio (\# Post-intevention samples assigned to class 1 / \# Post-intevention samples). Generally, the greater the value of parameter $d$, the higher the likelihood of a post-intervention being classified as class $1$. The variances are implicitly given by the fluctuations in the accuracies.}
  \label{fig:experiemts}
\end{figure}

Figure \ref{fig:experiemts} shows the results for different values of $d$, where the accuracy indicates the ratio of post-intervention samples classified as class $1$ compared to the total number of post-intervention samples. As the figure shows, when an optimal intervention is performed, larger $d$ corresponds to more samples being assigned to class $1$. On the other hand, when a naive intervention is performed, only approximately $68$\% of the post-intervention samples are assigned to class $1$, even for high values of $d$. If $d$ is small, the number of samples in each class is more balanced, whereas the majority of samples after optimal interventions can still be expected to be assigned to class $1$. Therefore, we could manipulate the data generation process such that the post-intervention observations had an arbitrarily high likelihood of being classified as class $1$.

\subsection{Real-world data}
In context of the experiments for this paper, it is difficult to conduct an evaluation using real-world data because of the necessary executions of the interventions and re-recording of the data. Therefore, generated DAGs provide a reasonable alternative to performing actual interventions. However, we wish to demonstrate a possible application using real-world data without actual interventions. Thus, we instead evaluated whether the suggested interventions would make logical sense.

For the demonstration, we took the popular Auto-MPG (miles per gallon) dataset from the University of California, Irvine (UCI) database. This dataset records the MPG consumption of $398$ automobiles. (Note that we omitted six samples with missing values.) The features are as follows: number of cylinders, weight, displacement, horsepower, acceleration, and MPG consumption. We determined the causal structure using LiNGAM and chose the MPG consumption as the target variable. As an exemplary application of our framework, we were interested in finding the necessary intervention such that a certain MPG consumption could be obtained. For the prediction, we chose linear regression as described in \eqref{eq:linearReg}.

Assuming that the inferred causal structure was valid, our framework suggested intervention on the number of cylinders, as this variable has the greatest causal effect on the prediction of the MPG consumption. For instance, the use of eight cylinders for an MPG consumption of $15$ was suggested, along with the use of six cylinders for a consumption of $21$ and the use of four cylinders for a consumption of $30$. These suggestions seem reasonable with respect to the recorded data and inferred causal structure. If we did not consider the causal influence and naively used \eqref{eq:closedForm} as intervention value, the results would make no logical sense; for example, a negative or an overly large number of cylinders would be suggested. 

Note that, for instance, in the case of predicting the number of cylinders, we might find an intervention to change the prediction of the numbers; however, any intervention on the other variables would not make physical sense in terms of changing the physical number.

\section{Conclusion}
In this work, we proposed a framework that allows identification of the dataset feature with the greatest causal influence on the prediction from a given prediction model, along with estimation of the optimal intervention on a feature such that a certain prediction can be expected. In particular, we provided an efficient algorithm that allows implementation of the framework for linear data. Via this framework, we want to contribute to a further and deeper understanding of prediction mechanisms from a causal perspective. The framework was successfully applied to artificial linear data and a possible application to a real-world dataset was also demonstrated. However, further evaluations using real-world data are an important topic for our future work. In addition, we are exploring modifications of the presented algorithms for non-linear data, which involve non-linear predictions and non-linear causal relationships. An extension for estimating the optimal interventions on multiple variables is another interesting aspect of this research. 

\begin{footnotesize}
\bibliographystyle{IEEEbib}
\bibliography{Literature}
\end{footnotesize}
\end{document}